\documentclass[11pt,a4paper]{llncs}
\usepackage{amsmath}
\usepackage{amssymb}
\usepackage{graphicx}
\usepackage{marvosym}
\usepackage{float} 
\usepackage{url}
\usepackage{fancyhdr}
\usepackage{authblk}

\usepackage{multirow}
\usepackage{subfigure}
\usepackage{hyperref}
\usepackage{geometry}
\newcommand{\keywords}[1]{\par\addvspace\baselineskip
\noindent\keywordname\enspace\ignorespaces#1}

\fancypagestyle{firstpage}{\fancyhf{}
}

\begin{document}


\title{\LARGE{Divergent Ensemble Networks: Enhancing Uncertainty Estimation with Shared Representations and Independent Branching}}


%
%
\author{\large{A. Chandorkar\textsuperscript{*} and A. Kharbanda\textsuperscript{*}}}

\newcommand{\suppressfootnotemark}{%
  \renewcommand{\thefootnote}{}%
  \footnotetext{\textsuperscript{*}These authors have contributed equally.}%
}

\newcommand{\resetfootnotemark}{%
  \renewcommand{\thefootnote}{\arabic{footnote}}%
  \setcounter{footnote}{0}%
}

\suppressfootnotemark
\resetfootnotemark



\institute{\large{Indian Institute of Technology, Ropar}}

%


%
%


\maketitle

\thispagestyle{firstpage}

\begin{abstract}
Ensemble learning has proven effective in improving predictive performance and estimating uncertainty in neural networks. However, conventional ensemble methods often suffer from redundant parameter usage and computational inefficiencies due to entirely independent network training. To address these challenges, we propose the Divergent Ensemble Network (DEN), a novel architecture that combines shared representation learning with independent branching. DEN employs a shared input layer to capture common features across all branches, followed by divergent, independently trainable layers that form an ensemble. This shared-to-branching structure reduces parameter redundancy while maintaining ensemble diversity, enabling efficient and scalable learning. 

%

\end{abstract}


\section{Introduction}

In controlled conditions and simulations, contemporary deep-learning networks have shown remarkable performance. However, when confronted with unexplored data, unforeseen fluctuations, or noisy inputs, its reliability frequently declines in real-world circumstances, potentially resulting in predictions that are not accurate. This raises concerns about the dependability of these networks in practical applications. \cite{Jiang20185541}. To address this challenge, leveraging uncertainty estimation was proposed as a solution to enhance their robustness and ensure real-world applicability. Neural networks struggle with quantifying predictive uncertainty and often generate overly confident predictions. Such overconfident incorrect predictions can lead to detrimental results\cite{lakshminarayanan2017simple}.

In deep learning, there are two primary types of uncertainties: \textit{aleatoric uncertainty} and \textit{epistemic uncertainty}. 

\begin{itemize}
    \item \textbf{Aleatoric uncertainty} originates from the inherent noise or randomness in the data. This type of uncertainty cannot be reduced by altering the model, as it is an intrinsic characteristic of the data itself \cite{Hullermeier2021}.
    \item \textbf{Epistemic uncertainty} arises from a lack of knowledge or information regarding the model. Unlike aleatoric uncertainty, epistemic uncertainty can be mitigated by improving the model design or gathering additional data \cite{Hullermeier2021}.
\end{itemize}

For \textbf{aleatoric uncertainty}, one approach to quantifying it is through the entropy of the predictive distribution. This allows us to measure the uncertainty inherent in the data \cite{NEURIPS2018_a981f2b7}.

On the other hand, \textbf{epistemic uncertainty} is typically quantified by assessing the variability between different models, which reflects how much our model predictions vary due to a lack of knowledge.

Several approaches have been proposed to handle these uncertainties:

\begin{itemize}
    \item \textbf{Bayesian Neural Networks (BNNs)} treat the network weights as random variables, thereby modeling uncertainty by learning distributions over the weights \cite{neal2012bayesian}.
    \item \textbf{Monte Carlo Dropout (MC Dropout)} approximates the posterior distribution by applying dropout during inference, enabling uncertainty estimation without explicitly Bayesian training \cite{gal2016dropout}.
    \item \textbf{Ensemble Methods}, such as Bagging, reduce uncertainty by combining predictions from multiple models. The diversity of the models helps in reducing the overall uncertainty \cite{breiman1996bagging}.
    \item \textbf{Bootstrapping} involves training multiple models on resampled subsets of the data. This method estimates uncertainty by evaluating the variance between models trained on different data subsets \cite{klas2011handling}.
    \item \textbf{Deep Ensembles} combine the predictions of multiple models to estimate uncertainty by analyzing the variance of their predictions \cite{ganaie2022ensemble}.
\end{itemize}


These methods offer distinct approaches to quantifying and reducing uncertainty in deep learning models, helping to enhance model reliability and performance. The ensemble's biggest challenge is reducing redundancy while preserving its diversity. While techniques like Monte Carlo Dropout (MC Dropout) approximate ensembles by stochastic regularization, they may compromise predictions' independence \cite{seoh2020qualitative}. Similarly, approaches like Batch Ensemble aim to reduce computation but need help with flexibility in capturing model variance. This trade-off between computational efficiency and one of the primary obstacles to ensemble learning is still ensemble diversity. To address these challenges, we propose the Divergent Ensemble Network (DEN), a novel architecture that combines the benefits of shared representation learning and ensemble diversity. DEN features a shared input layer that captures a common representation across all ensemble members, followed by independent branching layers that train separately to maintain prediction diversity. This architecture balances computational efficiency with predictive robustness, making it suitable for scenarios where both accuracy and uncertainty estimation are critical. 


In this study, we assess DEN across various tasks, emphasizing its ability to estimate uncertainty while preserving predictive accuracy. Our findings indicate that DEN surpasses conventional ensembles in computational efficiency and achieves similar or superior uncertainty estimation. Additionally, its scalable nature allows for adaptability in a wide range of applications, including those that require real-time predictions. The remainder of this paper is structured as follows: Section 2 provides an overview of the DEN architecture and its training approach. Section 3 showcases the findings, while Section 4 explores the applications and constraints, and Section 5 offers concluding remarks.

\section{Architecture}
\subsection{Problem setup and High-level summary}

We assume that the training dataset $\mathcal{D}$ consists of $N$ Independent and Identically Distributed (i.i.d) data points.
For classification problems, the label is assumed to be one of $K$ classes, that is $y \in \{1, \dots, K\}$. For regression problems, the label is assumed to be real-valued, that is $y \in \mathbb{R}$. 

Given the input features $x$, we use a neural network to model the probabilistic predictive distribution $p_\theta(y|x)$ over the labels, where $\theta$ are the parameters of the neural network.

The Divergent Ensemble Network (DEN) uses a shared-to-branching architecture for ensemble learning. A common input layer processes shared features, followed by a single shared representation layer that extracts fundamental features. After this, the network diverges into multiple independent branches, each trained with distinct parameters to have independent predictions. These branches operate on the shared features and produce independent outputs. This model thus provides independent predictions while also giving faster predictions.

\begin{figure}[ht]
    \centering
    \includegraphics[width=1\linewidth]{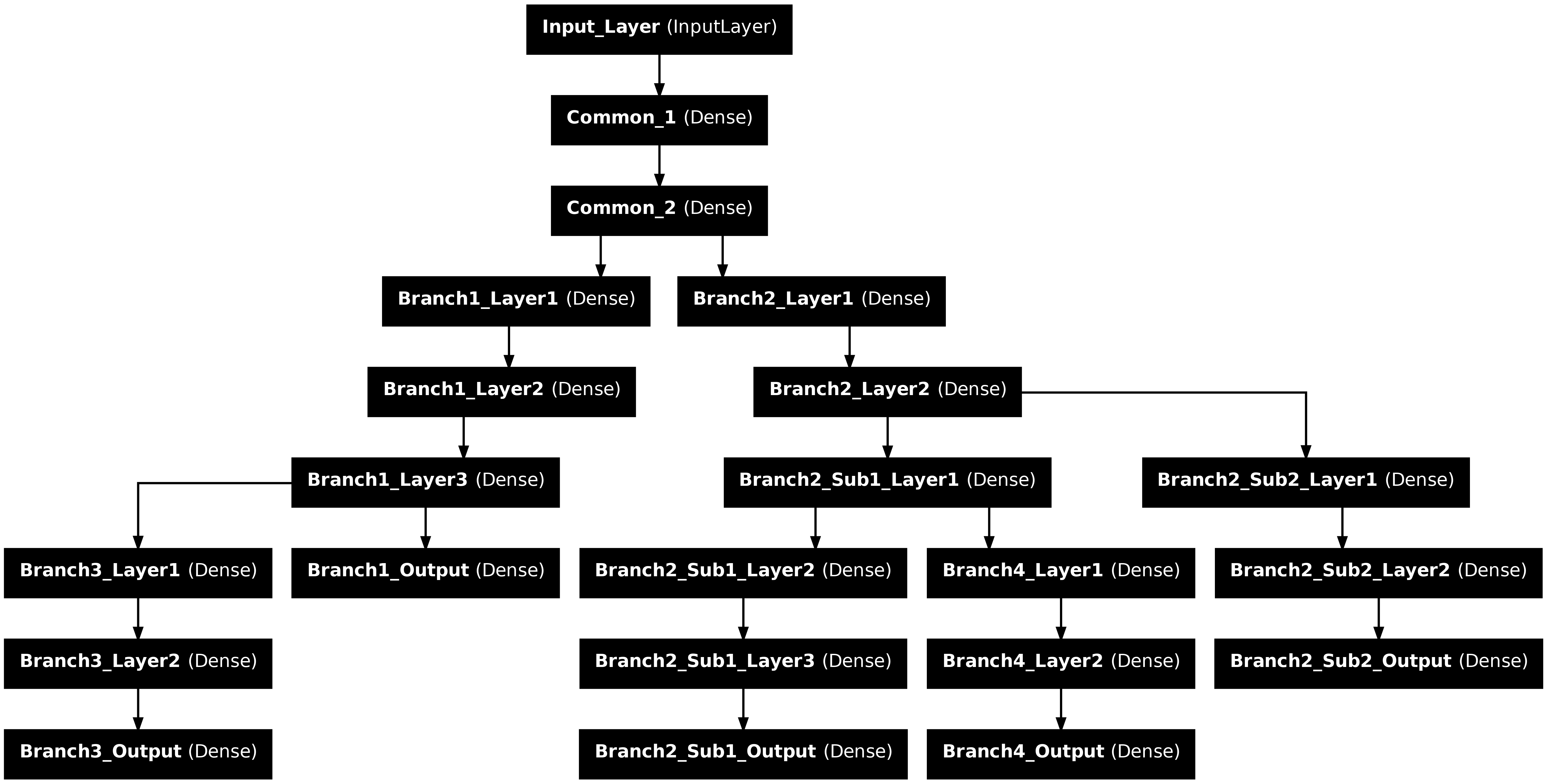}
    \caption{Visual representation of the proposed neural network architecture. The number of branches and the nodes in each branch are hyper-parameters and varies from problem to problem.}
    \label{fig:network_architecture}
\end{figure}

The shared layer in DEN ensures efficient parameter usage by processing input features once, reducing redundancy. Meanwhile, the divergent branches maintain the independence necessary for effective ensemble learning. By decoupling shared representation from individual predictions, DEN achieves a unique balance between computational efficiency and predictive variance.

\subsection{Shared Input and Common Representation Layer}
The architecture is divided into several independent networks after the shared layers. Each branch processes the shared representation further, utilizing distinct weights and biases to enable independent predictions. These branches act as independent ensemble members while benefiting from the standard input processing. Each branch consists of multiple fully connected layers, with the activation functions being themselves considered as a hyperparameter. Dropout layers are also incorporated, as they can improve accuracy by reducing overfitting.

However, increasing the number of branches introduces trade-offs in terms of computational resources. While adding more branches can potentially improve model performance by enhancing ensemble diversity, it also requires more memory and computational power, as each additional branch adds extra parameters and layers to the model. This results in higher processing time during both training and inference stages, which can be particularly costly in resource-constrained environments. Therefore, a balance must be struck between the number of branches and available computational resources to ensure efficient use of the system.

\subsection{Divergent Branches for Independent Predictions}

The architecture is divided into several independent networks after the shared layers. Each branch processes the shared representation further, utilizing distinct weights and biases to enable independent predictions. These branches act as independent ensemble members while benefiting from the common input processing. Each branch consists of multiple fully connected layers, with the activation functions being considered as a hyperparameter. Dropout layers are also incorporated, as they can improve accuracy by reducing overfitting.

However, increasing the number of branches introduces trade-offs in terms of computational resources. While adding more branches can potentially improve model performance by enhancing ensemble diversity, it also requires more memory and computational power, as each additional branch adds extra parameters and layers to the model. This results in higher processing time during both training and inference stages, which can be particularly costly in resource-constrained environments. Therefore, a balance must be struck between the number of branches and available computational resources to ensure efficient use of the system.

\subsection{Loss Function and Optimization}
Each branch is trained independently using a loss function specific to the task, and each branch gets to see the data in a loop so that the common layers are trained. For classification tasks, the softmax cross-entropy loss is typically used, and the mean squared error for regression tasks. Additionally, ensemble diversity is encouraged by training branches independently, ensuring variance in predictions and robust uncertainty estimation.

\section{Observations and Results}

\subsection{Dataset}
We first used the MNIST dataset to train our model. The MNIST dataset is a large, well-known database of handwritten digits comprising 70,000 images, of which 60,000 are used for training and 10,000 for testing. Each image is a grayscale representation of a digit (0–9) with a resolution of 28×28 pixels. As one of the benchmark datasets in machine learning and computer vision, MNIST provides a simple yet effective framework for evaluating classification algorithms. To further test our model's robustness, we utilized the NotMNIST dataset, which offers a similar structure but focuses on the classification of letters A through J in a variety of fonts. The dataset comprises 18,724 training images and 1,872 testing images, all grayscale and sized at 28×28 pixels.

\begin{figure}[H]
    \centering
    \includegraphics[width=0.45\linewidth]{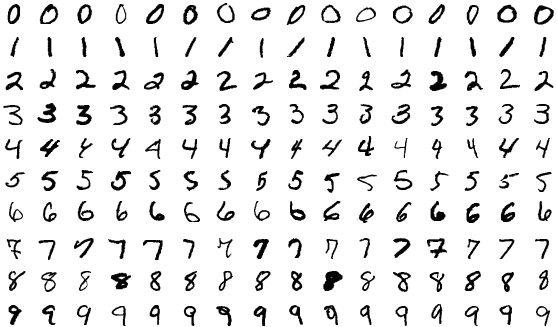}
    \hfill
    \includegraphics[width=0.45\linewidth]{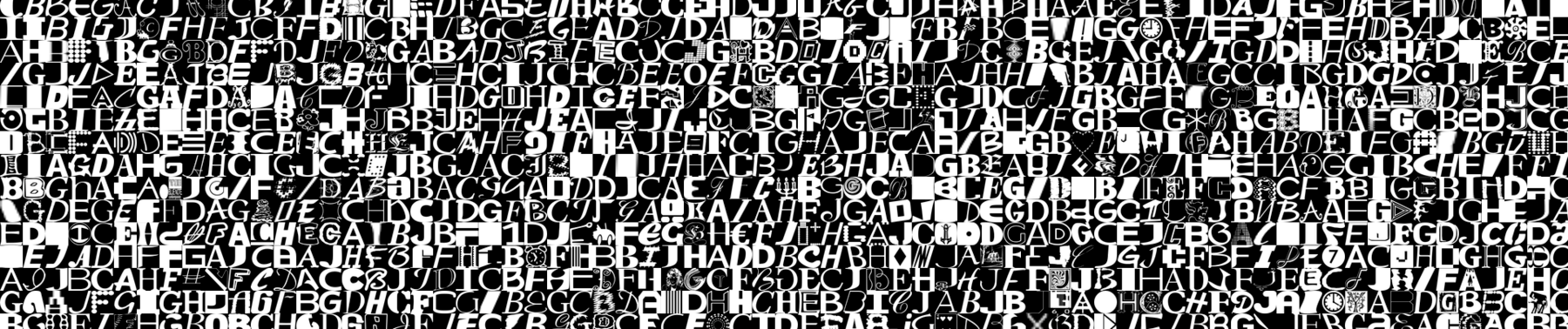}
    \caption{Comparison of the MNIST and NotMNIST Datasets.}
    \label{fig:datasets_comparison}
\end{figure}

After comparing the results with the MNIST dataset, we extended our evaluation to test the model's ability to quantify uncertainty by using it on the NotMNIST datasets. The NotMNIST dataset contains not only numbers but also letters, which provides a way to test how the model performs on unseen data. By analyzing the model's performance on these NotMNIST datasets, we aimed to evaluate how effectively it distinguishes between familiar (in-distribution) and unfamiliar (out-of-distribution) data. This approach has been widely practiced \cite{lakshminarayanan2017simple} involving unpredictable or previously unseen data. To evaluate the performance of our proposed model in uncertainty estimation on regression data we use a toy function.

\[
y = 10 \sin(x) + \epsilon,
\]

where \(\epsilon\) is a small noise term sampled from a normal distribution, \(\epsilon \sim \mathcal{N}(0, \sigma^2)\), added to introduce variability in the data. The input values, \(x\), are sampled from the range \([-3, 3]\) with a step size of 0.001. For \(x < 0\), the noise term is sampled with a standard deviation of \(\sigma_1 = 3\), and for \(x \geq 0\), the noise term has a standard deviation of \(\sigma_2 = 1\). 
The corresponding true values \(y_{\text{true}}\) are calculated without noise, following the equation

\[
y_{\text{true}} = 10 \sin(x).
\]

For uncertainty estimation, we extend the evaluation to include test data sampled from values outside the domain of the training dataset (e.g., $x \in [b+\delta, c]$ for some $\delta > 0$). This setup allows us to assess the model's ability to identify epistemic uncertainty, as these test points lie in regions not seen during training. The increased uncertainty in such out-of-distribution (OoD) regions provides a robust measure of the model's performance in handling unknown scenarios.

We compare the results of our model with other standard approaches such as deep ensembles, Bootstrap and MC Dropout. By examining the predicted values, their confidence intervals, and uncertainty metrics.

\subsection{Uncertainty Evaluation}
Overconfident predictions on unseen classes pose a challenge for the reliable deployment of deep learning models in real-world applications. Therefore, we expected the predictions to exhibit higher uncertainty for test data that was much different from the training data. To test if the proposed method possesses this desirable property, we trained a Multilayer Perceptron (MLP) on the standard MNIST train/test split using the proposed neural architecture. 

In addition to the regular test set with known classes, we also evaluated the model on a test set containing unknown classes. For this, we used the test split of the NotMNIST dataset, where the images have the same size as in the MNIST dataset, but the labels are alphabets instead of digits. Though we do not have access to the true conditional probabilities, we expected the predictions to be closer to a uniform value on unseen classes compared to the known classes, where the predictive probabilities should concentrate on the true targets. We evaluated the entropy of the predictive distribution and used this to assess the quality of the uncertainty estimates.

 We trained a Multilayer Perceptron (MLP) using the proposed neural architecture on a toy regression problem with Mean Squared Error (MSE) as the loss function to evaluate whether the proposed method demonstrates this desirable property on regression as well.

The training data consisted of input values sampled from a specific domain, $x \in [a, b]$, and corresponding noisy targets generated from the function $y = 10 \sin(x) + \epsilon$, where $\epsilon \sim \mathcal{N}(0, \sigma^2)$. 

After training, we tested the model on two distinct sets of data:  

1. \textbf{In-Domain Test Set:} Samples from the same range as the training domain ($x \in [a, b]$).  

2. \textbf{Out-of-Domain Test Set:} Samples drawn from a different range not seen during training ($x \in [b+\delta, c]$, for some $\delta > 0$).  

We hypothesized that the model’s predictions on the out-of-domain test set would reflect higher uncertainty compared to the in-domain test set, where the model had been trained to approximate the function. Since the true conditional variance of the target is known for the toy function, this setup allows us to qualitatively and quantitatively assess the model's ability to identify epistemic uncertainty.

To measure the uncertainty, we analyzed the entropy of the predictive distribution for each test point. For out-of-domain samples, we expect the predictions to reflect a more uniform distribution, indicating a lack of confidence. 

where \( \hat{y}_i \) is the predicted probability or confidence score for each sample.

\[
\text{Variance} = \frac{1}{n} \sum_{i=1}^{n} (\hat{y}_i - \bar{\hat{y}})^2
\]
where \( \hat{y}_i \) is the predicted value for each sample, and \( \bar{\hat{y}} \) is the mean of the predicted values.

\[
H(p) = - \sum_{i=1}^{n} p(x_i) \log p(x_i)
\]
where \( p(x_i) \) is the probability of class \( x_i \), and the sum is taken over all possible classes. Entropy measures the uncertainty of the predicted probability distribution.

\vspace{0.5cm}

For regression tasks, we evaluate performance using MSE, MAE, R², and average inference time. 
\[
\text{MSE} = \frac{1}{n} \sum_{i=1}^{n} (y_i - \hat{y}_i)^2
\]
where \( y_i \) is the true value and \( \hat{y}_i \) is the predicted value.

\[
\text{MAE} = \frac{1}{n} \sum_{i=1}^{n} |y_i - \hat{y}_i|
\]
where \( y_i \) is the true value and \( \hat{y}_i \) is the predicted value.

\[
R^2 = 1 - \frac{\sum_{i=1}^{n} (y_i - \hat{y}_i)^2}{\sum_{i=1}^{n} (y_i - \bar{y})^2}
\]
where \( y_i \) is the true value, \( \hat{y}_i \) is the predicted value, and \( \bar{y} \) is the mean of the true values.

As shown in Figure ~\ref{fig:regression_plots}, while the MSE, MAE, and R² scores remain relatively consistent, there is a significant reduction in inference time.

\subsection{Results}
\begin{table}[h!]
\centering
\caption{Evaluation Results (MNIST) (Classification)}
\label{tab:mnist}
\begin{tabular}{|l|c|c|c|}
\hline
\textbf{Model} & \textbf{Ensemble Accuracy (\%)} & \textbf{Single Model Accuracy (\%)} & \textbf{Average Inference Time (s)} \\
\hline
Ensemble & 98.56 & 97.78 & 0.263453 \\
MC Dropout & 97.33 & 97.67 & 0.158070 \\
Bootstrap & 97.67 & 96.67 & 0.277438 \\
\textbf{DEN (Ours)} & \textbf{98.78} & \textbf{98.44} & \textbf{0.056009} \\
\hline
\end{tabular}
\end{table}

\begin{table}[h!]
\centering
\caption{Evaluation Results (NotMNIST) (Classification)}
\label{tab:notmnist}
\begin{tabular}{|l|c|c|c|c|c|}
\hline
\multirow{1}{*}{\textbf{Model}} & \textbf{Accuracy} & \textbf{Average Confidence} & \textbf{Average Variance} & \textbf{Average Entropy} & \textbf{Inference Time (s)} \\
\hline
Ensemble & 10.22 & 0.6754 & 0.0406 & 0.6707 & 0.228371 \\
MC Dropout & 9.78 & 0.7545 & 0.0053 & 0.5987 & 0.164834 \\
Bootstrap & 10.78 & 0.6691 & 0.0416 & 0.6853 & 0.229866 \\
\textbf{DEN (Ours)} & 10.78 & 0.7019 & 0.0375 & 0.6125 & \textbf{0.045510} \\
\hline
\end{tabular}
\end{table}

\begin{table}[h!]
\centering
\caption{Evaluation Results (Regression)}
\label{tab:regression}
\begin{tabular}{|l|c|c|c|c|}
\hline
\multirow{1}{*}{\textbf{Model}} & \textbf{MSE} & \textbf{MAE} & \textbf{R²} & \textbf{Inference Time (s)} \\
\hline
Ensemble     & 5.1846 & 1.6353 & 0.9088 & 0.298662 \\
MC Dropout   & 5.4889 & 1.7045 & 0.9034 & 0.159630 \\
Bootstrap    & 5.2421 & 1.6383 & 0.9078 & 0.308347 \\
\textbf{DEN (Ours)}   & 5.3483 & 1.6703 & 0.9059 & \textbf{0.065887} \\
\hline
\end{tabular}
\end{table}

For the MNIST dataset, our model outperforms the existing Ensemble, MC Dropout, and Bootstrap methods by about 6x, and about 4x in the case of the NotMNIST dataset. All of the methods show low entropy as expected. However, MC Dropout seems to give high confidence predictions for some of the test examples, even for unseen classes. The results are shown in Tables~\ref{tab:mnist} and~\ref{tab:notmnist}. Such overconfident wrong predictions can be problematic in practice when tested on a mixture of known and unknown classes. DEN, Bootstrap, and Ensemble produce higher uncertainty on unseen classes, which is a desirable feature for reliable model deployment in real-world applications.

For regression on toy function our model outperforms the Ensemble and bootstrap by a factor of about 5 and MC Dropout by a factor of about 2.5. All the methods have about the same MSE, MAE and R² Score. The result is shown in Table~\ref{tab:regression}.

\section{Applications}
Uncertainty estimation methods have shown great potential in domains like computer vision and medical imaging \cite{rawat2017harnessing}. Deep ensembles effectively detect epistemic uncertainties, such as out-of-distribution (OoD) data and adversarial samples, enhancing neural network robustness \cite{rawat2017harnessing}. Integrating ensembles with probabilistic embeddings further improves uncertainty quantification, essential for real-world tasks \cite{scott2019stochastic}.

In continual learning, ensembles address uncertainty by enabling robust learning in dynamic environments \cite{shi2019probabilistic}. In medical imaging, they enhance confidence calibration and prediction reliability \cite{mehrtash2020confidence}. Divergent Ensemble Networks (DENs), with their multi-branch architecture, provide efficient uncertainty estimation and are faster at inference, making them ideal for real-time applications like robotics, manufacturing testing, and more.

\section{Conclusion}
This model achieved significant results on standard datasets for uncertainty estimation. The proposed architecture requires less space and computation time compared to traditional ensemble architectures. While Monte Carlo methods are more space-efficient than the proposed model, Bootstrap, ensembles, and Monte Carlo approaches require multiple inferences, increasing the time needed to generate results. In contrast, the proposed model is well-suited for time-sensitive and real-time applications.

\subsection{Limitations of DEN}
One of the current limitations of the Divergent Ensemble Networks (DEN) is its dependence on the choice of shared representations. The effectiveness of these representations is crucial for the overall model performance and may require substantial hyperparameter tuning. Furthermore, as the number of branches increases, computational resources required for both training and inference also increase. This can pose scalability challenges, particularly when dealing with very large ensembles. These factors may limit DEN's application to resource-constrained environments or larger datasets without further optimization.

\subsection{Future Work}
In the future, the proposed model should be implemented on hardware platforms to evaluate its performance in real-world scenarios, particularly for Internet of Things (IoT) devices and other real-time applications. The hardware implementation can reveal challenges related to computational overhead and integration with existing IoT ecosystems. Additionally, a scalability analysis for the proposed model needs to be conducted to better understand its performance as the number of branches increases. Optimizations aimed at reducing the computational burden and improving resource efficiency will be crucial for extending the applicability of DEN to larger, more complex tasks.

\subsection{Broader Impact}
DEN has the potential to make a positive societal impact by reducing biases in AI systems. By leveraging uncertainty estimation, DEN can enhance the fairness and adaptability of models, making them more robust to changes in input data and ensuring that decisions are made with higher confidence. Furthermore, DEN can contribute to ethical AI deployment by providing transparency into model predictions and supporting decision-making processes that are more explainable and reliable.



\bibliographystyle{plain}
\bibliography{refrences} 

\appendix
\section{Code Availability}
The code used for the experiments in this paper is publicly available on GitHub: \url{https://github.com/Arker123/Divergent-Ensemble-Networks}.

\section{Hardware Setup}
We ran the model on an Intel i5 12th gen processor, utilizing only a single core for performance comparison purposes. We found an approximate 6x improvement in execution time compared to other models (see Table~\ref{tab:mnist}), making it suitable for real-time predictions. We tested the model on 9000 examples from each class and evaluated the uncertainty on out-of-distribution examples from unseen classes for classification.





\section{Results}
The following plots showcase the robustness of our model (DEN). As shown, DEN excels in the inference time metric compared to all other models, while maintaining comparable performance across the other metrics.

\begin{figure}[ht]
    \centering
    \includegraphics[width=0.24\linewidth]{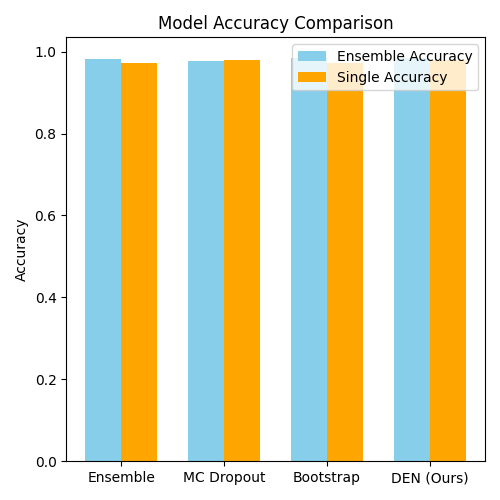}
    \includegraphics[width=0.24\linewidth]{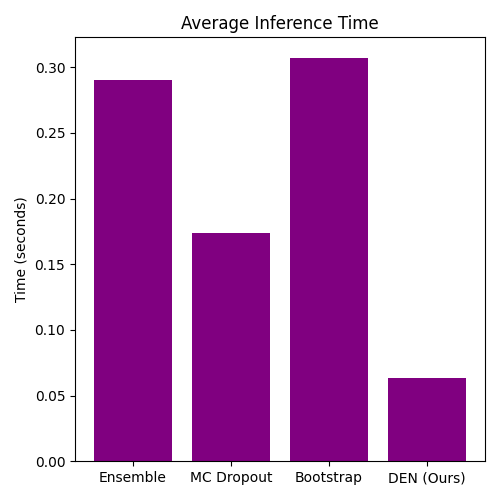}
    \includegraphics[width=0.24\linewidth]{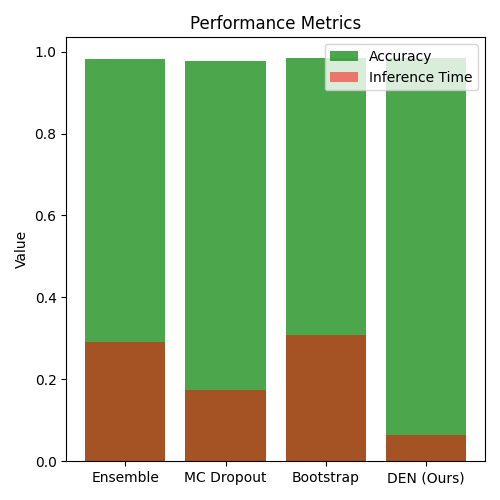}
    \caption{Classification performance metrics: (a) Model Accuracy Comparison between Ensemble and Single Model Approaches, (b) Average Inference Time per Model, (c) Performance Metrics: Accuracy and Inference Time for Different Models}
    \label{fig:classification_plots}
\end{figure}

\begin{figure}[ht]
    \centering
    \includegraphics[width=0.24\linewidth]{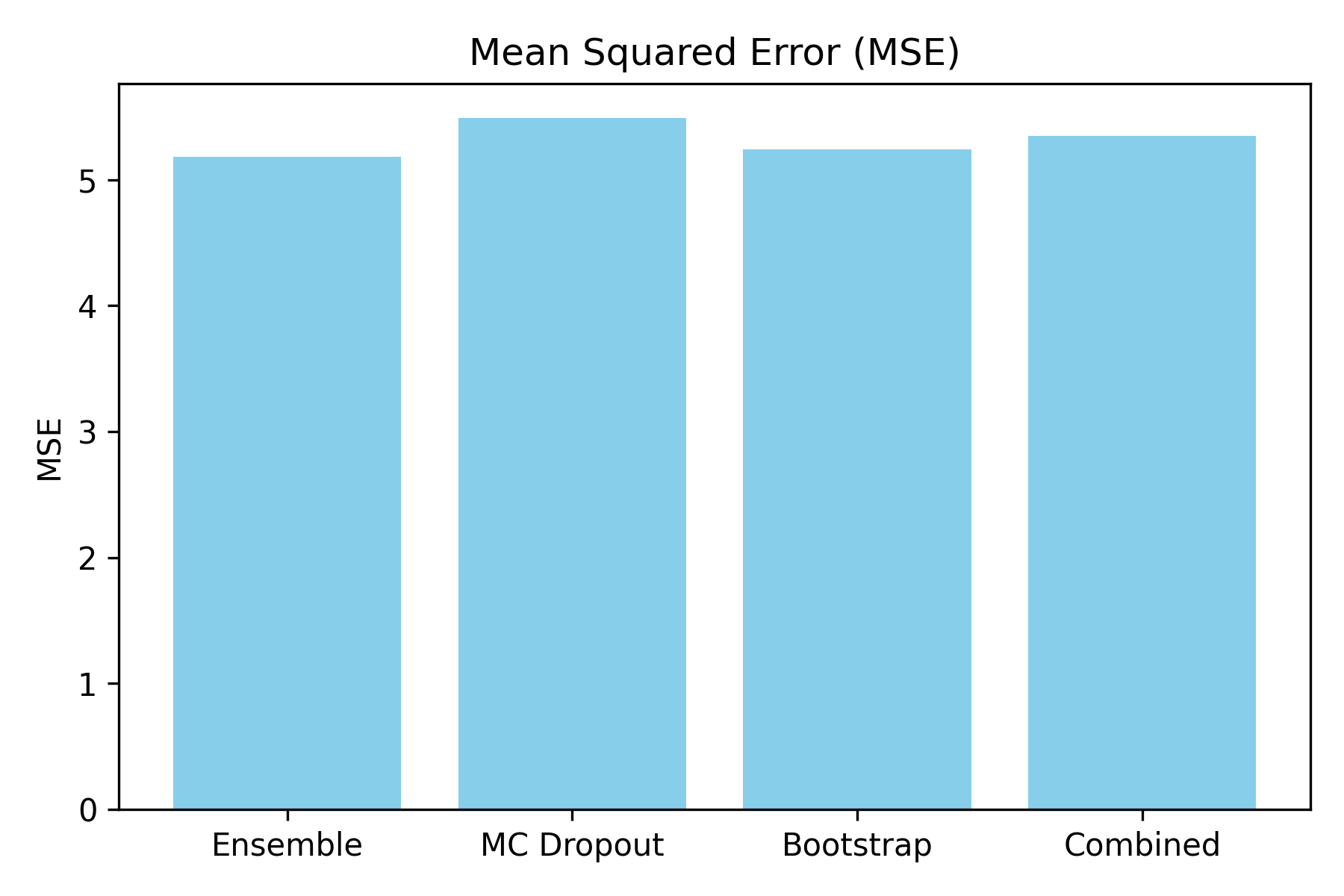}
    \includegraphics[width=0.24\linewidth]{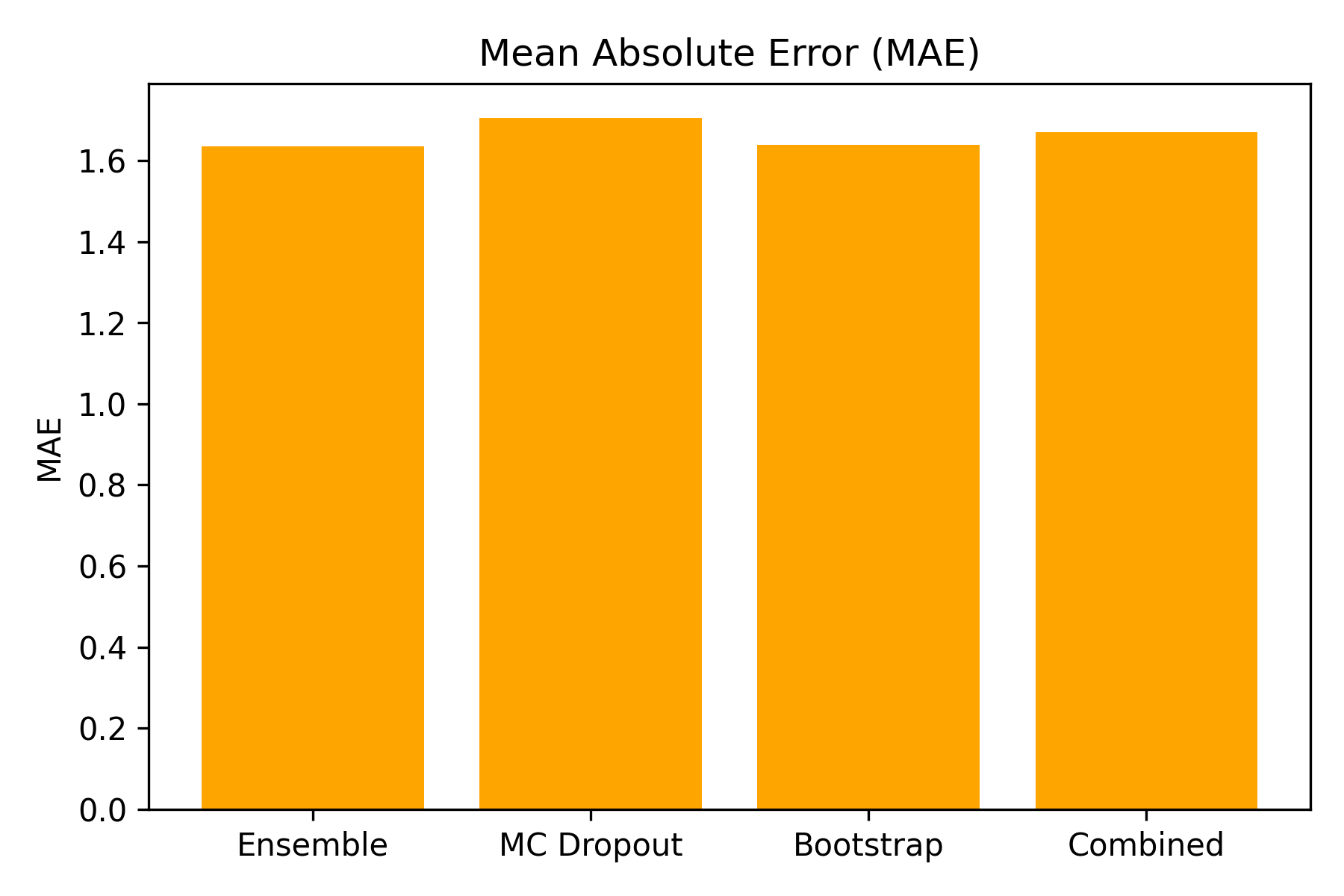}
    \includegraphics[width=0.24\linewidth]{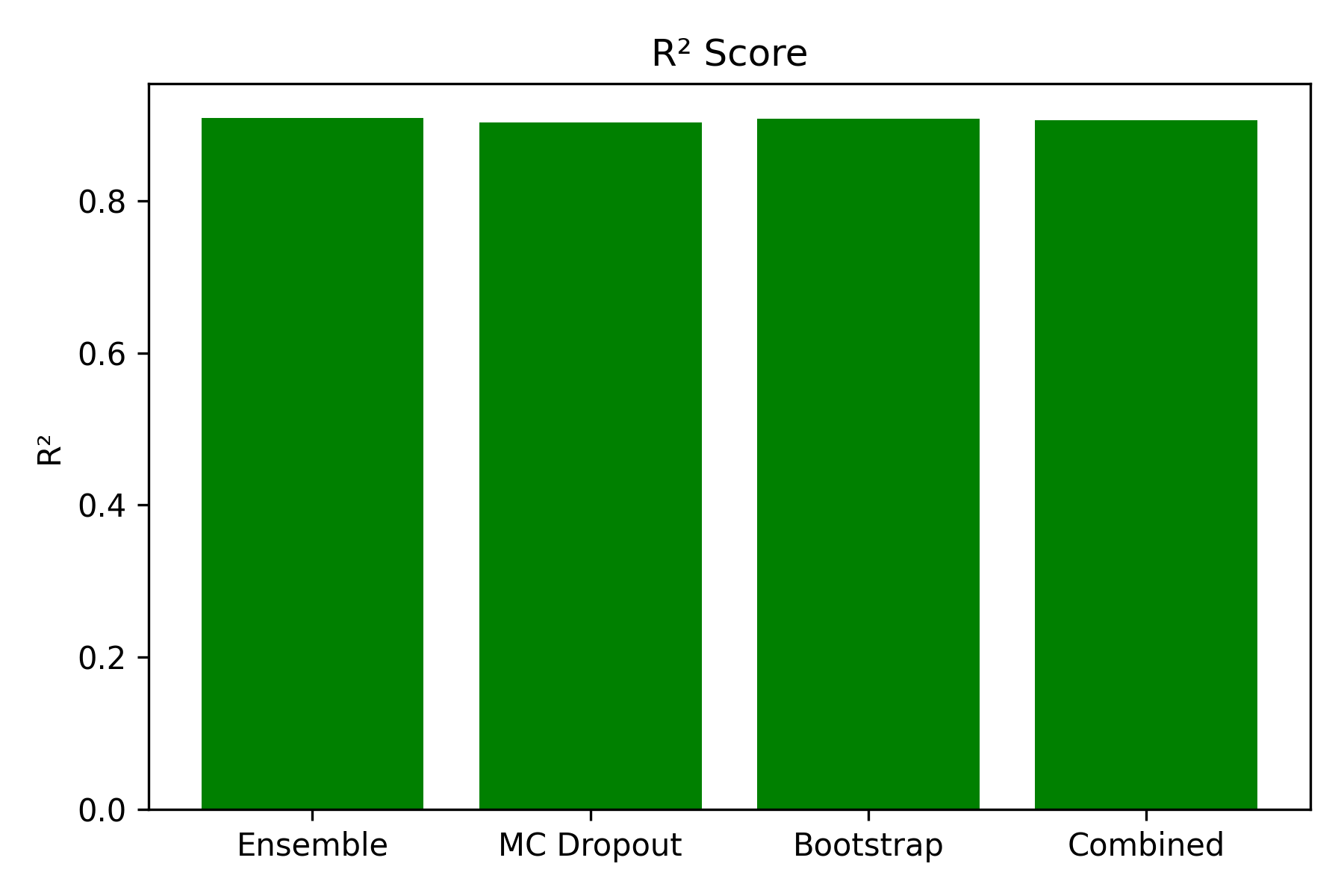}
    \includegraphics[width=0.24\linewidth]{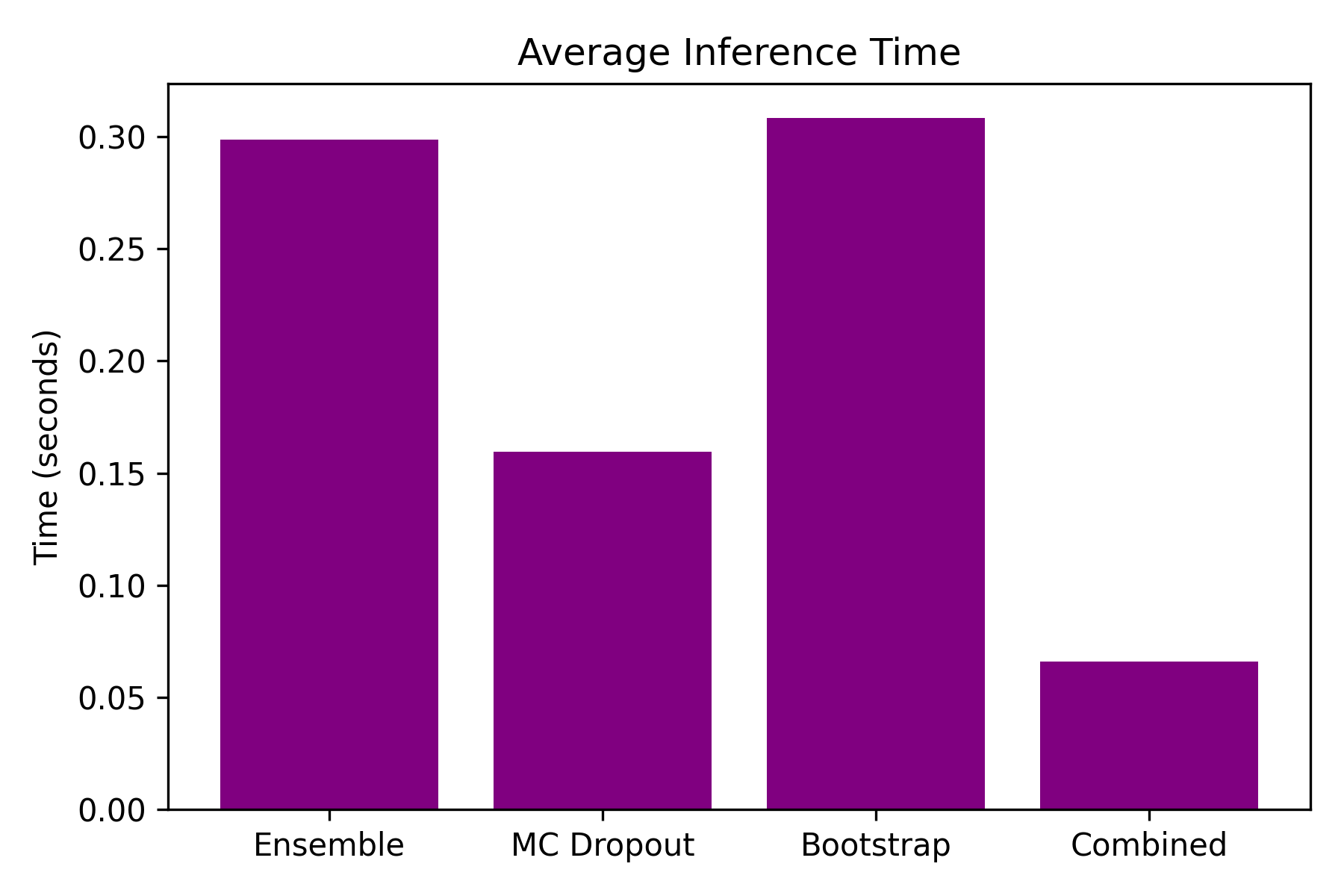}
    \caption{Regression performance metrics: (a) Mean Squared Error (MSE), (b) Mean Absolute Error (MAE), (c) R² Score, and (d) Average Inference Time.}
    \label{fig:regression_plots}
\end{figure}

\vspace{5 cm}
{\noindent  \large \textbf{Authors}} \vspace{0.5cm} \\
\noindent {\bf Advait Chandorkar} is currently studying mechanical engineering at IIT Ropar. His research interests include Robotics, Evolutionary Game Theory, and Computational Mechanics.

\noindent {\bf Aranav Kharbanda} is currently studying Computer Science Engineering at IIT Ropar. His research interest includes Robotics, Reinforcement Learning, and Machine Learning.

\end{document}